\documentclass[letterpaper, 10 pt, conference]{ieeeconf}  

\IEEEoverridecommandlockouts                              

\usepackage{cite}
\usepackage{amsmath,amssymb,amsfonts}
\usepackage{algorithmic}
\usepackage{graphicx}
\usepackage{textcomp}
\def\BibTeX{{\rm B\kern-.05em{\sc i\kern-.025em b}\kern-.08em
    T\kern-.1667em\lower.7ex\hbox{E}\kern-.125emX}}
\usepackage{dsfont}
\usepackage{paralist}
\usepackage[font=footnotesize]{subcaption}
\usepackage{cite}
\usepackage{tabularx}
\usepackage{cancel}
\usepackage{bm}
\usepackage{url}
\usepackage{mathtools}
\usepackage{textcomp}
\usepackage{nccmath}
\usepackage{xcolor}
\usepackage{multirow}
\usepackage{hhline}
\usepackage{capt-of,etoolbox}
\usepackage{graphicx}
\usepackage{etoolbox}
\usepackage{makecell}

\usepackage{todonotes}
\usepackage{booktabs}
\usepackage[table]{xcolor}

\title{\LARGE \bf
Extended Model-Based Learned Inertial Odometry
}

\author{Sasanka Kuruppu Arachchige$^{1}$ and Joni-Kristian Kämäräinen$^{1}$
\thanks{*This work was supported by the European Commission’s
Marie Skłodowska-Curie Action (MSCA) Project RAICAM
(GA101072634) 2025.}
\thanks{$^{1}$Authors are with the Computing Sciences department at Tampere University, Finland
       {\tt\small sasa.kuruppuarachchi@gmail.com}
       {\tt\small joni.kamarainen@tuni.fi}
       }%
}

\begin{document}

\graphicspath{{Pictures/}}


\newcommand{\insertfig}{\includegraphics[width=\textwidth]{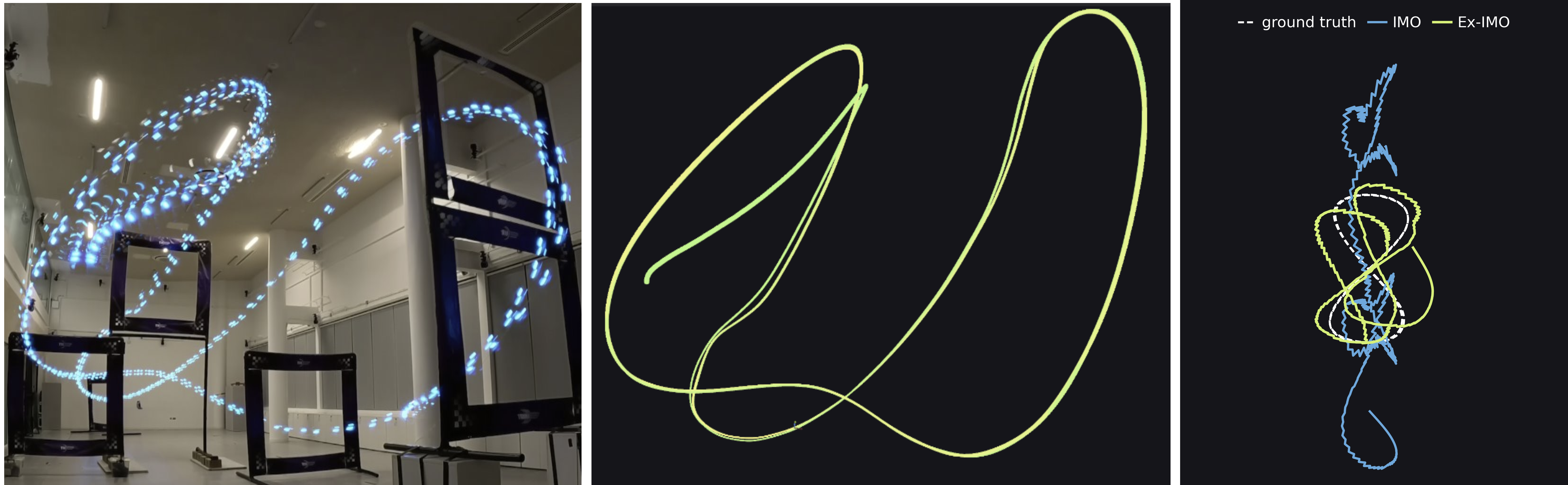}\captionof{figure}{By using the dynamic model of the drone to derive body forces and torques and feeding them into the learned model, we are able to accurately track the trajectory of an autonomous racing drone using an IMU as the only sensor modality. \textbf{Left:} The autonomous drone \cite{iitdataset} performs an agile maneuver. \textbf{Middle:} The trajectory estimated by our method. \textbf{Right:} Top view of a flight pattern never seen in training (public AI-IO dataset~\cite{cui2026ai}, the flight with the median improvement): ground truth (dashed), IMO~\cite{CioffiRal2023} (blue) and our method (green).}}\label{coverfig}

\makeatletter
\apptocmd{\@maketitle}{\centering\insertfig}{}{}
\makeatother

\maketitle

\thispagestyle{empty}
\pagestyle{empty}


\begin{abstract}

Inertial odometry is a compelling approach to state estimation for agile quadrotor flight due to its affordability, low weight, and robustness to perceptual degradation. However, relying only on integrated inertial measurements is impractical, as sensor errors and time-varying biases lead to significant pose drift. Although recent advances have enabled inertial odometry for drone racing, existing approaches show limited generalization to trajectories not seen during training. This work improves generalization by adopting a body-frame representation and incorporating body-frame torque dynamics into the learning pipeline. The learned model predicts short-horizon relative displacements, which are fused with IMU measurements in an Extended Kalman Filter. Experimental results show that the proposed method outperforms the previous state-of-the-art learned inertial odometry approach for quadrotor pose estimation on unseen trajectories. An additional evaluation on four public quadrotor datasets, with whole trajectory shapes held out of training, confirms a lower velocity error on every held-out set and shows that the torque input is most informative when it is derived from the rotor signals. Code, data pipeline and trained models are publicly available.

\end{abstract}


{\raggedright{\bfseries\textit{Multimedia Material}}--- Code, data pipeline and trained models:\\{\small\url{https://github.com/SasaKuruppuarachchi/Ex-IMO}}\par}
\section{INTRODUCTION}
Agile autonomous flying robots, such as quadrotors, are used in time-critical missions including search and rescue, aerial delivery, and reconnaissance \cite{watkins2020ten}. Accurate state estimation is essential for enabling this autonomy. Inertial odometry (IO) is a promising state-estimation approach in resource-constrained systems because of its low computational and storage requirements. Moreover, IO is not affected by common visual-odometry failure modes such as motion blur, low-texture environments, and high-dynamic-range scenes. These conditions are common during agile flight maneuvers, where dedicated visual processing is often required~\cite{kuruppu2024robust}. Although 6-DoF poses can be estimated by integrating inertial measurements, commodity IMUs suffer from bias drift, noise, scale-factor errors, misalignment, and vibration-induced inaccuracies, which cause long-term estimation drift \cite{yang2020online}. If reliable pose estimation can be achieved using only inertial sensing, robots can be made smaller and lighter, improving agility and operating range.
Recently, deep learning has enabled progress in inertial odometry for state estimation. For example, \cite{liu2020tlio} show that motion priors for pedestrians can be learned from IMU measurements, with accuracy comparable to recent visual-inertial odometry (VIO) methods.

Unlike pedestrian motion, quadrotor motion lacks a consistent prior that can be learned directly from IMU measurements. However, the quadrotor dynamics model can serve as a motion prior, which can then be corrected using onboard measurements (e.g., IMU data) and control outputs.
Cioffi~\textit{et al.}~\cite{CioffiRal2023} introduced a method that incorporates a quadrotor dynamics model and collective thrust to learn quadrotor translational motion for a given trajectory. The resulting relative position displacements are used to update an EKF propagated from IMU measurements. The method is accurate on known trajectories in the drone-racing context; however, its generalization to unseen trajectories remains unclear.

Our work adopts the general framework of Cioffi~\textit{et al.}~\cite{CioffiRal2023}. We introduce two main modifications to improve its performance:
%
\begin{itemize}
    \item We introduce body-frame torque dynamics into the learned inertial odometry pipeline, enabling the model to account for angular acceleration–induced effects that are otherwise absorbed as unmodeled noise.

    \item We reformulate the learning problem in the body frame, improving generalization and observability.

    \item We experimentally validate the model by integrating it into a state-of-the-art learning-based inertial odometry framework and comparing it with prior art, on Blackbird and DIDO and on four further public quadrotor datasets with held-out trajectory shapes.



\end{itemize}
The effectiveness of the proposed modifications is demonstrated on several widely used offline datasets, and the new model is deployed on a real drone.
To facilitate fair comparisons, the code, the data pipeline and the trained models are publicly available at \url{https://github.com/SasaKuruppuarachchi/Ex-IMO}.

\section{RELATED WORK}


Odometry, the process of estimating a robot's position and orientation over time from its internal sensor data, is important for drones, which often operate indoors, below the GPS canopy, or in urban canyons.

\textbf{Visual-Inertial Odometry} (VIO) is often the preferred choice, since accurate, low-drift pose estimation from visual input is essential for tasks ranging from stable hovering to fast maneuvers in complex environments. The main VIO approaches are filter-based and optimization-based methods.
Filter-based methods belong to the multistage constraint Kalman filter (MSCKF) family derived from~\cite{Mourikis-2007-icra}. Optimization-based methods, such as OKVIS~\cite{OKVIS} and Kimera-VIO~\cite{Kimera}, are often found to be more accurate than filtering-based methods. Recently, hybrid filtering-optimization approaches have reported superior accuracy by combining the two approaches~\cite{HybVIO}. 

\textbf{IMU-only odometry} is beneficial over VIO in certain scenarios due to its robustness to visual degradation, lower computational demand, and independence from lighting and environmental conditions. IMU odometry works by measuring acceleration and angular velocity at very high rates, allowing real-time motion estimation even when cameras fail or the environment is visually ambiguous (e.g., darkness, featureless walls, smoke). This makes IMU odometry especially valuable for high-dynamic maneuvers and visually challenging environments as a complementary subsystem or short-term fallback for VIO.

\textbf{Learning inertial odometry} can substantially improve IMU-only odometry by addressing its core limitations, most notably sensor drift, non-linearity, and bias sensitivity. 
Deep learning can be used to estimate velocity and displacement from IMU data. Methods like IONet~\cite{IONet}, A2DIO~\cite{A2DIO} and NILoc~\cite{NILoc} treat inertial navigation as a sequential task by predicting relative motion. These predictions are often integrated into EKF or optimization frameworks. To maintain consistency, many approaches transform IMU data into global coordinates such as a heading-agnostic coordinate frame (HACF)~\cite{RoNIN}. 
Most learning-based IO methods are designed for pedestrian datasets, making them less effective for UAVs with more complex motion dynamics.
DIDO~\cite{zhang2022dido} estimates the thrust of the UAV using tachometer data and compensates for unmodeled forces with a neural network. IMO~\cite{CioffiRal2023} extends this by integrating thrust and IMU data into an EKF but relies on additional control signals, which risk overfitting.
The starting point for this work is the IMO method by Cioffi~\textit{et al.}~\cite{CioffiRal2023}. We extend it with a richer dynamics model and replace the global-coordinate representation with a local body-frame representation, improving its suitability for diverse UAV maneuvers.



\section{METHODOLOGY}

\begin{figure}[thpb]
\setcounter{figure}{1}
      \centering
      
      \includegraphics[width=\textwidth/2]{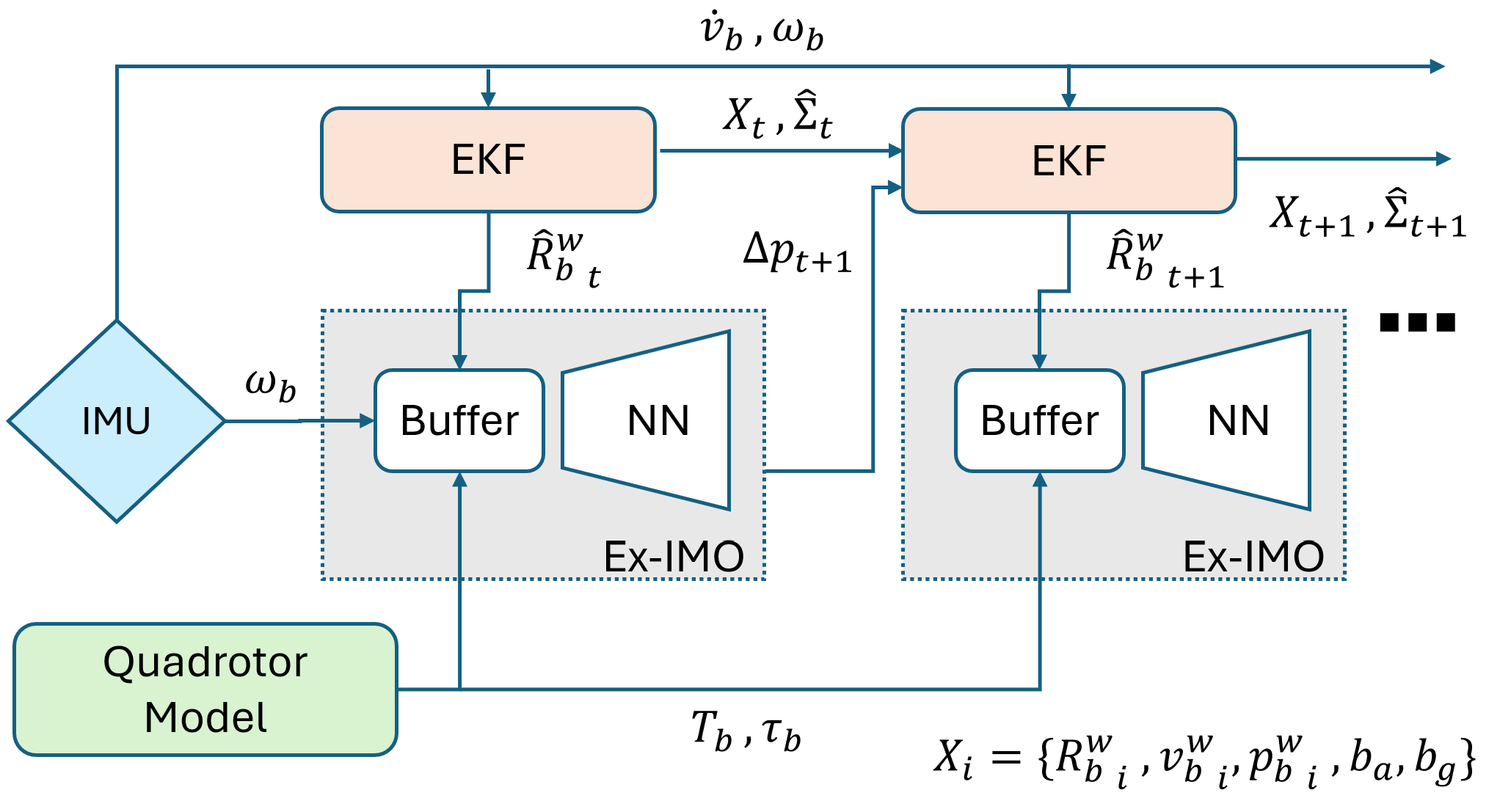}
      \caption{Block diagram of our system. The temporal convolution neural network (TCN) takes as input a buffer of collective thrust and gyroscope measurements and outputs relative 3-DoF positional displacements. These displacements are used to update an EKF, which is propagated using the IMU measurements}
      \label{Overview}
   \end{figure}

Our model is inspired by~\cite{CioffiRal2023} to which we introduce two important changes, change of the global coordinate frame to the drone/IMU body coordinate frame and a new dynamics model.
An overview of our system is shown in Fig.~\ref{Overview}. The learning-based element is a temporal convolutional network (TCN) which we train to regress 3-DoF relative displacements from a buffer of length $\Delta t$ seconds containing mass-normalized collective thrust and gyroscope measurements. These displacements represent the distance traveled by the quadrotor in the time interval $\Delta t$. We use the predictions of the network to update an Extended Kalman Filter (EKF), which is propagated using a kinematic motion model of the IMU.

\subsection{IMU Coordinate Frame}
\label{sec:frames}

\subsubsection{Canonical Body Coordinate Frame}
The IMU accelerometer measures the net force per unit mass in the body frame, including translational acceleration $\frac{\mathbf{F}_{b_i}}{m}$ caused by external forces and gravitational acceleration $\mathbf{g}^w$. The accelerometer reading at the timestamp $i$ is
\begin{equation}
    \mathbf{\dot{v}}^b_i = \frac{\mathbf{F}_{b_i}}{m} + \mathbf{R}_i^\top \mathbf{g}^w,
    \label{eq:body-frame-acc}
\end{equation}
where $\mathbf{R}_i^T$ is the drone attitude with respect to the world frame, and $\mathbf{g}^w$ is gravity in that global frame.

\subsubsection{Global Coordinate Frame}
If the drone pose is tracked in global coordinates, it can be convenient to convert the IMU measurements in the body frame (\ref{eq:body-frame-acc}) to measurements in the global or heading-agnostic coordinate frame (HACF) using an estimated rotation matrix $\hat{\mathbf{R}}_{b_i}^w$,
\begin{equation}
    \mathbf{\dot{v}}^G_i = \hat{\mathbf{R}}_{b_i}^w \cdot \left( \frac{\mathbf{F}_{b_i}}{m} \right) + \hat{\mathbf{R}}_{b_i}^w \mathbf{R}_i^\top \mathbf{g}^w.
    \label{eq:global-frame-acc}
\end{equation}

Although Equations (\ref{eq:body-frame-acc}) and (\ref{eq:global-frame-acc}) describe the same physical quantities, they encode attitude differently. In the body frame model in Eq.~\ref{eq:body-frame-acc}, the attitude affects only the gravity component, forming a linear relationship. In the global frame model in Eq.~\ref{eq:global-frame-acc}, attitude couples with drone motion, requiring more complex modeling and inducing rotation matrix estimation errors to a component which is free of them in the body frame model. This makes body-frame data more favorable for learning-based estimation tasks due to simpler attitude-force coupling.

\subsection{Learned Quadrotor Model}

We use the same notation as Cioffi~\textit{et~al.}~\cite{CioffiRal2023} to make it easier to compare the two models. The reference frame $\mathcal{W}$ is the fixed world frame with its z-axis aligned with gravity. The frame of the quadrotor body is denoted as $\mathcal{B}$ and the IMU frame coincides with it. We denote by a superscript the reference frame of quantities. For example, $\mathbf{p}^w_{b_i} \in \mathbb{R}^3$ is the position of the body frame at time $t_i$ in the world frame. For notational simplicity, the time instant is denoted just by the index $i$. Similarly, $\mathbf{v}^w_{b_i} \in \mathbb{R}^3$ is the linear velocity of the body frame,
$\boldsymbol{\omega}_{b_i}$ orientation velocity, and $\mathbf{R}^w_{b_i} \in \mathbb{R}^{3 \times 3}$ rotation. 

\subsubsection{Quadrotor Model}

The motion of the quadrotor is governed by the two equations~\cite{CioffiRal2023}:
\begin{align}
    \dot{\mathbf{p}}^w_{b_i} &= \mathbf{v}^w_{b_i}, \\
    \dot{\mathbf{v}}^w_{b_i} &= \mathbf{R}^w_{b_i} \cdot (\mathbf{T}_{b_i} + \mathbf{F}^b_{e_i}) + \mathbf{g}^w, \label{eq:accel_model}
\end{align}
where $\mathbf{T}_{b_i}$ is the mass-normalized collective thrust, and $\mathbf{F}^b_{e_i}$ is the external force acting on the quadrotor in the body frame. Since the external forces are unknown, we assume $\mathbf{F}^b_{e_i} \sim \mathcal{N}(\mathbf{0}, \Sigma)$.

More specifically, the control input to the system, $\mathbf{u} \in \mathbb{R}^4$, can be decomposed into the total thrust $u_1$ and the body torques $\boldsymbol{\tau} = [u_2, u_3, u_4]^T$, yielding two additional terms~\cite{pinto2021planning}:
\begin{align}
    \dot{\mathbf{R}^w_{b_i}} &= \mathbf{R}^w_{b_i} S(\boldsymbol{\omega}_{b_i})\\
    \mathbf{J}\dot{\boldsymbol{\omega}_{b_i}}&+S(\boldsymbol{\omega}_{b_i})\mathbf{J}\boldsymbol{\omega}_{b_i} = \boldsymbol{\tau}
\end{align}
where $\mathbf{J}$ represents the inertia tensor about the frame $\mathcal{B}$ and $S(\boldsymbol{\omega})$ is a map from $\mathbb{R}^3$ to the space of three-by-three 
skew-symmetric matrices $SO(3)$, 
\begin{equation}
\mathcal{S}(\boldsymbol{\omega}) = 
\begin{bmatrix}
0 & -\omega_3 & \omega_2 \\
\omega_3 & 0 & -\omega_1 \\
-\omega_2 & \omega_1 & 0
\end{bmatrix} . 
\end{equation}

By integrating Equation~\ref{eq:accel_model} over a time interval $[t_i, t_{i+1}]$, assuming constant thrust $\mathbf{T}_{b_i}$ and no external force, we get the relative displacement:
\begin{equation}
    \Delta \mathbf{p}_{i,i+1} = \mathbf{v}^w_{b_i} \cdot \Delta t + \frac{1}{2} \cdot \mathbf{g}^w \cdot \Delta t^2 + \frac{1}{2} \cdot \mathbf{R}^w_{b_i} \cdot \mathbf{T}_{b_i} \cdot \Delta t^2.
\end{equation}

Solving for the thrust yields
\begin{equation}
    \mathbf{T}_{b_i} = 2 \cdot (\mathbf{R}^{b_i}_w) \left( \frac{\Delta \mathbf{p}_{i,i+1}}{\Delta t^2} - \frac{\mathbf{v}^w_{b_i}}{\Delta t} - \mathbf{g}^w \right)
    \label{eq:thrust_id}
\end{equation}
and solving for the torque yields
\begin{equation}
    \boldsymbol{\tau}_{b_i} = \mathbf{J} \dot{\boldsymbol{\omega}}_{b_i} + S(\boldsymbol{\omega}_{b_i}) \mathbf{J} \boldsymbol{\omega}_{b_i} .  \label{eq:torque_model}
\end{equation}

\subsubsection{Thrust and Torque Inputs}
Eq.~(\ref{eq:thrust_id}) contains the displacement and the velocity that the filter is estimating, so it cannot provide the thrust input at inference time. We use it only as the fitting target of a per-platform rotor map. With $u_k$ the signal of rotor $k$ (rotor speed or motor command), the mass-normalized thrust is
\begin{equation}
    T_{b_i} = \sum_{k=1}^{4} f_{k,i}, \qquad f_{k,i} = a\,u_{k,i}^2 + b\,u_{k,i},
    \label{eq:thrust_map}
\end{equation}
where $a$ and $b$ are fitted by least squares on the training split. A motor delay and a battery-voltage scaling are added when the platform logs commands instead of rotor speeds. Mass and thrust coefficient are therefore not needed separately.

The inertia tensor $\mathbf{J}$ is rarely documented, so we normalize Eq.~(\ref{eq:torque_model}) by it,
\begin{equation}
    \mathbf{J}^{-1}\boldsymbol{\tau}_{b_i} = \dot{\boldsymbol{\omega}}_{b_i} + \mathbf{c} \odot \begin{bmatrix} \omega_2\omega_3 & \omega_3\omega_1 & \omega_1\omega_2 \end{bmatrix}^\top,
    \label{eq:torque_norm}
\end{equation}
so that only the inertia ratios $\mathbf{c} = [\tfrac{J_z - J_y}{J_x}, \tfrac{J_x - J_z}{J_y}, \tfrac{J_y - J_x}{J_z}]$ enter. The torque input can be obtained in two ways. The \textit{gyro-derived} torque evaluates Eq.~(\ref{eq:torque_norm}) with a causal derivative of the gyroscope. It is a function of the gyroscope history that the network already receives, so it adds structure to the input but no new information. The \textit{rotor-derived} torque $\hat{\boldsymbol{\tau}}_{b_i} = \mathbf{A}\mathbf{f}_i + \mathbf{d}$ is a linear allocation of the per-rotor thrusts $\mathbf{f}_i$ of Eq.~(\ref{eq:thrust_map}), with $\mathbf{A}$ and $\mathbf{d}$ fitted by least squares against the gyro-derived torque on the training split. It does not depend on the gyroscope, and the motor layout is learned rather than assumed. Section~\ref{sec:public} compares the two.

\subsubsection{Role of Torque Dynamics in Translational Motion}
In quadrotor systems, translational acceleration is determined not only by thrust magnitude but also by thrust direction, which depends on vehicle orientation. Consequently, rotational dynamics indirectly influence translational motion through the orientation-dependent thrust vector. During aggressive maneuvers, rapid angular accelerations produced by body torques can induce transient misalignment between thrust direction and instantaneous vehicle orientation. If torque dynamics are not explicitly modeled, these effects appear as unmodeled disturbances in the translational dynamics of Eq.~(\ref{eq:accel_model}). Learning-based inertial odometry methods must then implicitly compensate for these disturbances using data-driven priors, which can lead to trajectory-dependent representations and reduced generalization. By incorporating the torque dynamics described in Eq.~(\ref{eq:torque_model}), the proposed model exposes angular motion as a structured component of the system dynamics rather than treating it as latent noise. This allows the learning module to better separate translational motion induced by thrust from motion caused by rotational dynamics, improving robustness to aggressive maneuvers and generalization to unseen trajectories.
\subsubsection{Neural Net Model}
To enable a direct comparison with~\cite{CioffiRal2023}, we keep the remaining components unchanged and use the same Temporal Convolutional Network (TCN) architecture to learn positional displacements $\Delta \mathbf{p}_{i,j}$. TCNs have proven effective for temporal modeling and are more efficient to train and deploy than recurrent architectures.

The input to the TCN is a sequence of collective thrust and gyroscope measurements over the interval $\Delta t_{i,j} = t_j - t_i$. In IMO, these measurements are rotated to the world frame using ground truth (for training) or EKF-estimated orientations (for inference), and the network predicts the world-frame displacement $\mathbf{p}^w_{b_j} - \mathbf{p}^w_{b_i}$. In Ex-IMO, they stay in the body frame. The input at each step $k$ is
\begin{equation}
    \mathbf{x}_k = \left[\boldsymbol{\omega}_{b_k},\; \mathbf{T}_{b_k} + \mathbf{R}^{b_k}_w \mathbf{g}^w,\; \mathbf{J}^{-1}\boldsymbol{\tau}_{b_k}\right],
\end{equation}
so that, as in Eq.~(\ref{eq:body-frame-acc}), the attitude enters only through the gravity term. The target is the displacement expressed in the body frame at the start of the window, $\Delta \mathbf{p}^{b_i}_{i,j} = \mathbf{R}^{b_i}_w (\mathbf{p}^w_{b_j} - \mathbf{p}^w_{b_i})$, which does not depend on the absolute heading.
To improve generalization, we apply zero-mean Gaussian noise to orientation and gyroscope bias during training. The output of the network is the predicted relative displacement $\Delta \hat{\mathbf{p}}_{i,j}$. The mean squared error (MSE) loss is used to train the model:
\begin{equation}
    \mathcal{L}(\Delta \mathbf{p}, \Delta \hat{\mathbf{p}}) = \frac{1}{N} \sum_{k=1}^{N} \left\| \Delta \mathbf{p}_k - \Delta \hat{\mathbf{p}}_k \right\|^2,
\end{equation}
where $N$ is the number of training samples.

\subsubsection{Filter Update}
The EKF keeps a window of past states and uses the network output as a relative-position measurement between the states at $t_i$ and $t_j$. For IMO, the measurement model is $\mathbf{h} = \mathbf{p}^w_{b_j} - \mathbf{p}^w_{b_i}$. For the body-frame prediction of Ex-IMO, it becomes
\begin{equation}
    \mathbf{h} = \mathbf{R}^{w\top}_{b_i} \left(\mathbf{p}^w_{b_j} - \mathbf{p}^w_{b_i}\right),
\end{equation}
with Jacobians $\partial\mathbf{h}/\partial\mathbf{p}^w_{b_j} = -\partial\mathbf{h}/\partial\mathbf{p}^w_{b_i} = \mathbf{R}^{w\top}_{b_i}$ and, for the attitude error $\delta\boldsymbol{\theta}_i$ defined by $\mathbf{R}^w_{b_i} = \mathrm{Exp}(\delta\boldsymbol{\theta}_i)\hat{\mathbf{R}}^w_{b_i}$, $\partial\mathbf{h}/\partial\delta\boldsymbol{\theta}_i = \hat{\mathbf{R}}^{w\top}_{b_i} S(\mathbf{p}^w_{b_j} - \mathbf{p}^w_{b_i})$. Unlike the world-frame update, the body-frame update also corrects the attitude.

\subsubsection{Implementation}

We train our model on a system running Ubuntu 20.04 with an Intel Core i9 CPU and an Nvidia RTX 4000 GPU. For real experiments, the trained model was deployed on an NVIDIA Jetson TX2 onboard the quadrotor. IMU and thrust measurements are sampled at 100 Hz, and the TCN receives a 0.5 s buffer as input. Network inference and EKF updates run at 20 Hz. With a history of 10 states, the system runs at $\sim$180 Hz on the Jetson TX2.

\section{EXPERIMENTS} 

\subsection{Error metrics}
Our model is evaluated against the original learning-based Inertial Model Odometry (IMO) in~\cite{CioffiRal2023}. We report the two commonly used estimation error metrics, the \textit{absolute trajectory error} (ATE) and the \textit{relative error} (RE), calculated between a ground truth trajectory $\mathbf{X}_{gt}$ and the aligned estimated trajectory $\hat{\mathbf{X}}'$. The alignment of the estimated trajectory, $\hat{\mathbf{X}}' = \hbox{align}(\hat{\mathbf{X}}, \mathbf{X}_{gt})$, is computed as a similarity transformation~\cite{Zhang-2018-iros}. For alignment and calculation of error metrics, we used the Python toolbox provided by the authors of~\cite{Zhang-2018-iros}. For all experiments, we report the absolute and relative translation errors $ATE_T$ and $RE_T$. The relative error is less sensitive to failures in single trajectory segments and can therefore provide more detailed information about accuracy. The segment length parameter affects the relative error computation, and we rely on the default values of the provided evaluation code. To simplify the analysis, we do not report the orientation errors since we found them to accumulate into the translation error.

\subsection{Datasets}

The \textbf{Blackbird} dataset provides rotor speed measurements, IMU data, and synthetic images for aggressive quadrotor flights. We compute mass-normalized collective thrust using rotor speeds and evaluate on the five dataset trajectories: \textit{Clover}, \textit{Egg}, \textit{Half Moon}, \textit{Star}, and \textit{Winter}, with peak speeds of 4–8 m/s. The dataset contains precise drone trajectories obtained by a Vicon motion capture system, and the sequences are trimmed to contain only agile flying.
The \textbf{DIDO} dataset provides IMU measurements and ground-truth poses for dynamic quadrotor trajectories. We evaluate five trajectory types: \textit{Circle}, \textit{Eight}, \textit{Updown Circle}, \textit{Updown Eight}, and \textit{Random}, with peak speeds ranging from 0.7 to 4.0 m/s. This dataset is more representative of real quadrotor flight than Blackbird because it contains longer human-piloted sequences with takeoff, hover, and landing.
%
%
The datasets are split into 70\% for training, 15\% for validation, and 15\% for testing.
Both datasets also include additional sequences that are excluded from training and used only for evaluation as ``unseen'' trajectories to assess generalization.

\begin{figure*}[h]
    \centering
    \includegraphics[width=\textwidth]{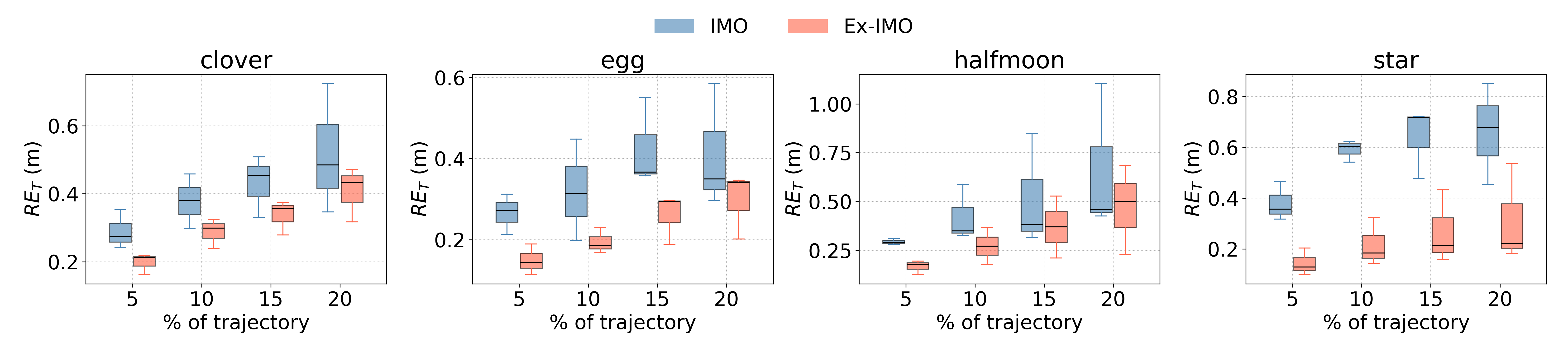}
    \caption{Blackbird dataset evaluation for the unseen cases. Relative translation errors achieved by IMO, and our extended IMO (Ex-IMO). The quantity on the x-axis is the percentage traveled corresponding to 5, 10, 15, and 20\% of the total distance traveled.}
    \label{fig:bb_box_rpe}
\end{figure*}

\begin{figure*}[h]
    \centering
    \includegraphics[width=\textwidth]{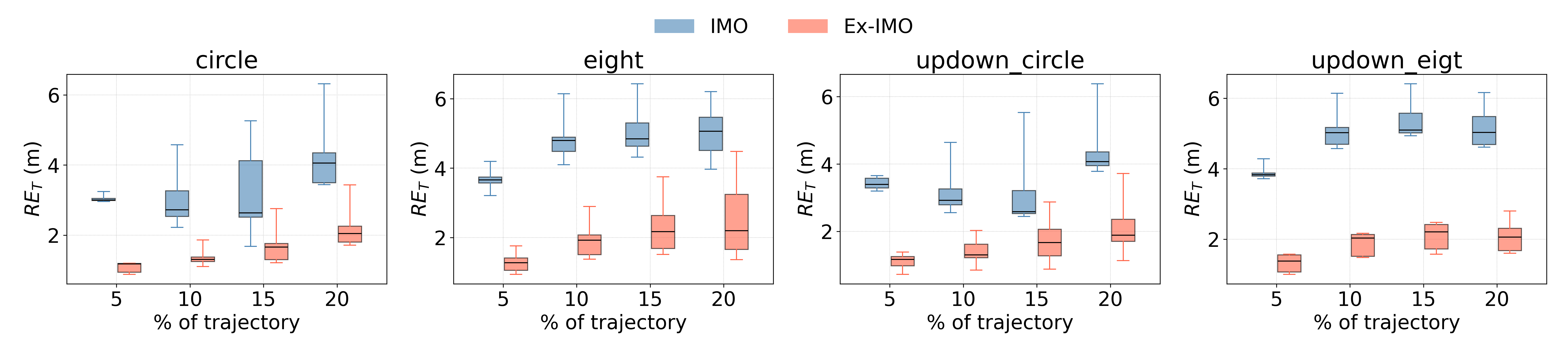}
    \caption{DIDO dataset evaluation for the unseen cases. Relative translation errors achieved by IMO and our extended IMO (Ex-IMO). The quantity on the x-axis is the percentage traveled, corresponding to 5, 10, 15, and 20\% of the total distance traveled.}
    \label{fig:dido_box_rpe}
\end{figure*}

\subsection{Results}

The Blackbird results are shown in Table~\ref{tab:blackbird_results}. The ``seen'' results are computed on test trajectories that were included during training, whereas ``unseen'' results are computed on hold-out trajectories not used in training. Figure~\ref{fig:bb_box_rpe} shows that our method performs consistently better on unseen trajectories. On average, our method improves $ATE_T$ by 28\% in the unseen setting. Although the original IMO performs better on seen trajectories, our method generalizes better to unseen data, thanks to the proposed extended dynamics model.


\begin{table}[h]
\centering
\caption{Blackbird dataset evaluation. ATE and RE values in meters. In bold is the best value.}
\label{tab:blackbird_results}
\resizebox{1.0\linewidth}{!}{%
\begin{tabular}{l c c c c}
\toprule
Method & \multicolumn{2}{c}{IMO~\cite{CioffiRal2023}} & \multicolumn{2}{c}{Our}\\
Trajectory & $ATE_T$ & $RE_T$ & $ATE_T$ & $RE_T$\\
\midrule
\multicolumn{5}{c}{\cellcolor{gray!25}Seen trajectories} \\
Clover     & \textbf{0.381}  & \textbf{0.681}  & 0.902  & 2.014 \\
Egg        & 1.153  & \textbf{0.828}  & \textbf{1.007}  & 1.398 \\
HalfMoon   & \textbf{0.761}  & \textbf{0.240}  & 1.496  & 1.008 \\
Star       & \textbf{2.130}  & \textbf{3.066}  & 2.713  & 3.796 \\
Winter     & \textbf{0.219}  & \textbf{0.206}  & 0.801  & 0.991 \\
Avg. & \textbf{0.929} & \textbf{1.004} & 1.384 & 1.841 \\
\midrule
\multicolumn{5}{c}{\cellcolor{gray!25}Unseen trajectories} \\
Ampersand  & 17.664 & 10.039 & \textbf{9.517} & \textbf{6.887} \\
Sid        & 9.967  & 6.992  & \textbf{7.693}  & \textbf{5.009} \\
Oval       & 5.670  & 2.863  & \textbf{5.012}  & \textbf{2.397} \\
Sphinx     & 5.629  & 5.395  & \textbf{4.891}  & \textbf{4.598} \\
BentDice   & 6.143  & 4.519  & \textbf{5.301}  & \textbf{3.913} \\
Avg.       & 9.015  & 5.962 & \textbf{6.483} & \textbf{4.561} \\
\hline
\end{tabular}}
\end{table}

\begin{table}[ht]
\centering
\caption{DIDO dataset evaluation. ATE and RE values in meters. In bold is the best value.}
\label{tab:dido_results}
\resizebox{1.0\linewidth}{!}{%
\begin{tabular}{l c c c c}
\toprule
Method & \multicolumn{2}{c}{IMO~\cite{CioffiRal2023}} & \multicolumn{2}{c}{Our}\\
Trajectory & $ATE_T$ & $RE_T$ & $ATE_T$ & $RE_T$\\
\midrule
\multicolumn{5}{c}{\cellcolor{gray!25}Seen trajectories} \\
Circle          & 5.451 &  2.720 & \textbf{4.045}  & \textbf{1.325} \\
Eight           & 4.916 & 3.816 & \textbf{3.511}  & \textbf{2.023} \\
Updown Circle   & 5.069 & 3.131 & \textbf{2.735}  & \textbf{1.110} \\
Avg.            & 5.145 & 3.222 & \textbf{3.430} & \textbf{1.486} \\
\midrule
\multicolumn{5}{c}{\cellcolor{gray!25}Unseen trajectories} \\
Updown Eight    & 6.2312 & 4.990  & \textbf{3.677}  & \textbf{2.337} \\
Random         & 12.1707 &  3.944 & \textbf{7.901}  & \textbf{1.173} \\
Avg.           & 9.201  & 4.467 & \textbf{5.789} & \textbf{1.755} \\
\hline
\end{tabular}}
\end{table}

\begin{table}[ht]
\centering
\caption{Factorial ablation on Blackbird unseen trajectories using the four-way protocol of global/body-frame prediction with and without torque inputs. Values are $ATE_T$ in meters. In bold is the best value.}
\label{tab:blackbird_ablation}
\resizebox{1.0\linewidth}{!}{%
\begin{tabular}{l c c c c}
\toprule
Seq. & Global & Global + torques & Body & Body + torques \\
\midrule
Ampersand & 17.664 & 14.487 & \textbf{9.245} & 9.517 \\
Sid & 9.967 & 8.984 & 7.678 & \textbf{7.693} \\
Oval & 5.670 & 5.571 & 5.078 & \textbf{5.012} \\
Sphinx & 5.629 & 5.784 & 5.266 & \textbf{4.891} \\
BentDice & 6.143 & 6.626 & 5.784 & \textbf{5.301} \\
Avg. & 9.015 & 8.290 & 6.610 & \textbf{6.483} \\
\hline
\end{tabular}}
\end{table}

Table~\ref{tab:blackbird_ablation} reports a factorial ablation on the Blackbird unseen split with four variants: the original global-frame formulation, global-frame prediction augmented with torques, body-frame prediction without torques, and the full body-frame-plus-torques model. This protocol isolates the effect of each design choice and clarifies how they interact.

The ablation shows that the body-frame reformulation is the dominant source of improvement. Averaged over the unseen Blackbird trajectories, switching from the global frame to the body frame reduces $ATE_T$ from 9.015 to 6.610~m even without torque inputs, whereas adding torques in the global frame yields a smaller reduction from 9.015 to 8.290~m. Adding torques on top of the body-frame model further improves the average $ATE_T$ from 6.610 to 6.483~m, and gives the best result on four of the five sequences. The only exception is \textit{Ampersand}, where the body-frame-only variant is slightly better than the full model. Overall, the results indicate that the two modifications are complementary rather than redundant: the body-frame representation provides most of the gain, while the torque terms provide an additional refinement once the prediction target is expressed in the body frame.

The DIDO results are shown in Table~\ref{tab:dido_results}. DIDO is more representative of real agile quadrotor flight because it includes longer sequences and behaviors such as takeoff, hovering, and landing. As shown in Table~\ref{tab:dido_results}, our method performs significantly better in both seen and unseen cases. In unseen sequences, our method improves $ATE_T$ by 37\% and $RE_T$ by 61\%. For example, on the \textit{Random} trajectory sequence, IMO performance deteriorates substantially due to multiple hover maneuvers.

More detailed error and variance trends are shown in Figures~\ref{fig:bb_box_rpe} and \ref{fig:dido_box_rpe}. Together with the factorial ablation in Table~\ref{tab:blackbird_ablation}, these results show that the largest improvement comes from expressing the learning target in the body frame, while the torque dynamics provide a smaller but still consistent gain in the stronger body-frame formulation. This is particularly evident on novel trajectories not present in the training set.

\subsection{Evaluation on Public Quadrotor Datasets}
\label{sec:public}

To test whether these findings carry over to other platforms, we repeated the comparison with the released implementation on four further public quadrotor datasets, under a stricter protocol: several seeds, model selection on validation data only, and whole trajectory shapes held out of training.

\textbf{Datasets.} \textbf{AI-IO}~\cite{cui2026ai} (34 min, rotor speeds) contains manual flights and patterned flights (\textit{circle}, \textit{eight}, \textit{updown circle}, \textit{updown eight}, \textit{random}, at two speeds) that are never used in training. \textbf{NanoBench}~\cite{ullah2026nanobench} (57 min) was recorded with a 41~g nano-quadrotor that logs motor PWM commands. \textbf{TII-RATM}~\cite{iitdataset} (33 min) is the autonomous racing drone of Fig.~1 and logs motor commands and battery voltage. \textbf{NeuroBEM}~\cite{bauersfeld2021neurobem} (75 min, rotor speeds) contains aggressive flights of a quadrotor of about 0.76~kg. Duplicate recordings, samples with interpolated ground truth and ground-truth gaps were removed, and all data were resampled to 100~Hz, so the numbers are not directly comparable with those reported on the original files. The thrust map of Eq.~(\ref{eq:thrust_map}) explains 84--95\% of the variance of the inverse-dynamics thrust on the validation flights of AI-IO, TII-RATM and NeuroBEM, and 15\% on NanoBench, whose gentle flights have nearly constant thrust.

\textbf{Protocol.} Each model is trained with three seeds (Adam, learning rate $10^{-4}$, early stopping on validation loss). The EKF noise parameters, the measurement update rate and the torque source of Ex-IMO are chosen on validation flights only (105 filter settings per model), and the test flights are evaluated once. For the held-out experiment, each fold removes one group of trajectory shapes from training, validation and the physics fits: the \textit{ellipse}, \textit{lemniscate} and \textit{race track} tracks of TII-RATM, the crossing (\textit{figure-8}, \textit{star}, \textit{trefoil}) and round (\textit{circle}, \textit{oval}, \textit{helix}) shapes of NanoBench, and the \textit{circles} and \textit{lemniscate and CPC} flights of NeuroBEM. Together with AI-IO's patterned flights, this gives eight held-out sets. Since IMU-only odometry has metric scale, we align the trajectories with an SE(3) transformation and report $ATE_T$ together with the RMSE of the filter's velocity estimate, the quantity the network learns. Differences are tested with a paired t-test across seeds.

\begin{table}[ht]
\centering
\caption{Public datasets, trajectories of the training shapes (test split) and AI-IO's patterned flights. Velocity RMSE in m/s, $ATE_T$ in meters, mean of 3 seeds. In bold is the best value; $^{*}$: different from IMO with $p<0.05$.}
\label{tab:public_seen}
\resizebox{1.0\linewidth}{!}{%
\begin{tabular}{l c c c c c c}
\toprule
Method & \multicolumn{2}{c}{IMO~\cite{CioffiRal2023}} & \multicolumn{2}{c}{Our, gyro $\boldsymbol{\tau}$} & \multicolumn{2}{c}{Our, rotor $\boldsymbol{\tau}$}\\
Dataset & vel. & $ATE_T$ & vel. & $ATE_T$ & vel. & $ATE_T$\\
\midrule
\multicolumn{7}{c}{\cellcolor{gray!25}Seen trajectory shapes} \\
TII-RATM  & \textbf{0.516} & 1.302 & 0.773$^{*}$ & 1.428 & 0.721$^{*}$ & \textbf{1.153} \\
NanoBench & 0.479 & 1.756 & 0.458$^{*}$ & 1.801 & \textbf{0.378}$^{*}$ & \textbf{1.551}$^{*}$ \\
NeuroBEM  & 0.895 & 2.946 & 1.012$^{*}$ & 3.252 & \textbf{0.838} & \textbf{2.725} \\
AI-IO     & 1.384 & 4.322 & 1.420 & 5.244$^{*}$ & \textbf{0.519}$^{*}$ & \textbf{1.206}$^{*}$ \\
\midrule
\multicolumn{7}{c}{\cellcolor{gray!25}Unseen trajectories} \\
AI-IO patterned & 1.082 & 14.57 & 1.114 & 5.957$^{*}$ & \textbf{0.461}$^{*}$ & \textbf{2.897}$^{*}$ \\
\hline
\end{tabular}}
\end{table}

\begin{table}[ht]
\centering
\caption{Public datasets, held-out trajectory shapes: each row is a fold whose shapes were excluded from training, validation and the physics fits. Velocity RMSE in m/s, $ATE_T$ and $RE_T$ in meters, mean of 3 seeds. In bold is the best value; $^{*}$: $p<0.05$.}
\label{tab:public_heldout}
\resizebox{1.0\linewidth}{!}{%
\begin{tabular}{l c c c c c c}
\toprule
Method & \multicolumn{3}{c}{IMO~\cite{CioffiRal2023}} & \multicolumn{3}{c}{Our, rotor $\boldsymbol{\tau}$}\\
Held-out shapes & vel. & $ATE_T$ & $RE_T$ & vel. & $ATE_T$ & $RE_T$\\
\midrule
\multicolumn{7}{c}{\cellcolor{gray!25}NeuroBEM} \\
Circles          & 0.777 & 2.389 & 2.260 & \textbf{0.522}$^{*}$ & \textbf{1.205}$^{*}$ & \textbf{1.275}$^{*}$ \\
Lemniscate, CPC  & 1.111 & 4.114 & 4.370 & \textbf{0.718}$^{*}$ & \textbf{2.759}$^{*}$ & \textbf{2.821}$^{*}$ \\
\midrule
\multicolumn{7}{c}{\cellcolor{gray!25}NanoBench} \\
Figure-8, star, trefoil & 0.609 & 1.928 & 2.023 & \textbf{0.569} & \textbf{1.806} & \textbf{1.841} \\
Circle, oval, helix     & 0.557 & \textbf{1.325} & \textbf{1.436} & \textbf{0.470}$^{*}$ & 1.634 & 1.668 \\
\midrule
\multicolumn{7}{c}{\cellcolor{gray!25}TII-RATM} \\
Ellipse     & 5.728 & \textbf{4.965} & \textbf{6.823} & \textbf{2.185}$^{*}$ & 8.436$^{*}$ & 8.172$^{*}$ \\
Lemniscate  & 6.547 & 10.43 & 8.855 & \textbf{1.836}$^{*}$ & \textbf{9.613} & \textbf{6.736}$^{*}$ \\
Race track  & 8.456 & \textbf{16.89} & \textbf{14.93} & \textbf{4.808}$^{*}$ & 20.00 & 17.40 \\
\hline
\end{tabular}}
\end{table}

\begin{figure*}[t]
    \centering
    \includegraphics[width=\textwidth]{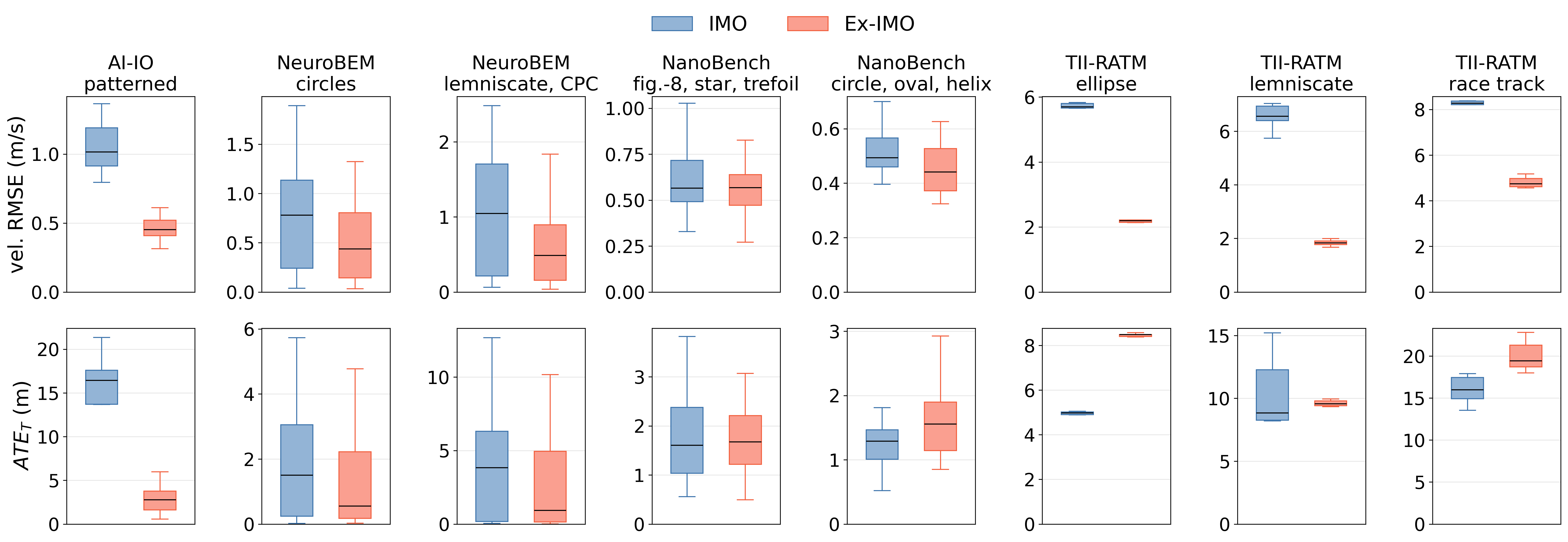}
    \caption{Public datasets, held-out trajectory shapes. Per-flight velocity RMSE (top) and $ATE_T$ (bottom) achieved by IMO and our extended IMO (Ex-IMO, rotor-derived torque), each flight averaged over 3 seeds. AI-IO's patterned flights are never used in training; every other panel is a fold whose shapes were excluded from training.}
    \label{fig:public_box}
\end{figure*}

\textbf{Results on seen trajectory shapes.} Table~\ref{tab:public_seen} compares IMO with the two torque sources of Ex-IMO. With the gyro-derived torque, Ex-IMO is not better than IMO on the training shapes: its velocity error is lower only on NanoBench and higher on TII-RATM and NeuroBEM. On AI-IO's patterned flights, however, it reduces $ATE_T$ from 14.57 to 5.96~m. Since the gyro-derived torque adds no information, this gain comes from the body-frame formulation, consistent with the Blackbird ablation in Table~\ref{tab:blackbird_ablation}. With the rotor-derived torque, which the validation data selected on all four datasets, Ex-IMO has the lowest error on three of the four datasets and reduces velocity RMSE and $ATE_T$ on AI-IO by 63\% and 72\%. On TII-RATM, IMO keeps the lowest velocity error, while Ex-IMO has the lower $ATE_T$.

\textbf{Results on held-out shapes.} On trajectory shapes that were never seen in training (Table~\ref{tab:public_heldout} and Fig.~\ref{fig:public_box}), Ex-IMO lowers the velocity error in all eight held-out sets, seven of them with $p<0.05$, by up to 72\%. The position error is lower in five of the eight sets, significantly on AI-IO ($-80\%$) and on both NeuroBEM folds ($-50\%$ and $-33\%$). NeuroBEM, with more than 100 training segments per fold, is the clearest case: Ex-IMO is better on every metric. TII-RATM has only six training flights per fold, and neither model generalizes across its tracks. There, Ex-IMO's velocity error is much lower, but its $ATE_T$ is higher on the \textit{ellipse} and \textit{race track} folds.

\textbf{Why the rotor-derived torque helps.} A linear regression from the rotor-derived torque alone explains 94\% and 71\% of the variance of the longitudinal and lateral body-frame displacement on AI-IO's validation flights, and 80\% and 78\% on NeuroBEM, while the gyro-derived torque explains almost none. Shuffling the rotor-derived torque between windows of the same flight, or delaying it by 1~s, removes the gain. The rotor differentials therefore carry information about the velocity, not only about the angular acceleration. A plausible explanation is that the rotors trim against aerodynamic moments that grow with airspeed~\cite{bauersfeld2021neurobem}, but we have not verified this. The torque term of Eq.~(\ref{eq:torque_model}) is thus most useful when it is measured through the rotors rather than reconstructed from the gyroscope.

\textbf{Heading dependence.} The world-frame formulation can memorize the headings of the training tracks. On TII-RATM, rotating the validation flights by $90^\circ$ in yaw raises the velocity error of IMO's network from 0.34 to 5.8~m/s, whereas the body-frame networks are unaffected. This supports the analysis of Section~\ref{sec:frames}.

\section{DISCUSSION}
The experimental results indicate that the proposed Ex-IMO framework consistently outperforms the original IMO method on the unseen trajectories of Blackbird and DIDO, while achieving comparable or slightly lower accuracy on trajectories observed during training. The factorial ablation further clarifies that this improvement is driven primarily by the body-frame formulation, with torque dynamics contributing an additional gain once the prediction problem is expressed in body coordinates. This behavior suggests a bias--variance trade-off introduced by the more structured representation. While the original IMO formulation can implicitly overfit to motion patterns present in the training dataset, the body-frame target and explicit modeling of rotational dynamics constrain the learned representation to follow more physically meaningful relationships.

The slightly weaker performance on some seen trajectories likely reflects this trade-off rather than a flaw in the proposed priors. In our view, the gap is due to a combination of stronger regularization and limited model capacity, rather than the priors being overly restrictive.

The evaluation on four public datasets refines this picture. The body-frame formulation removes the dependence on the absolute heading, which is what limits IMO on flight patterns with new headings. The value of the torque input depends on how it is obtained. Reconstructed from the gyroscope, it adds structure but no information, and it gives at most a small refinement (Table~\ref{tab:blackbird_ablation}) or none (Table~\ref{tab:public_seen}). Measured through the rotors, it carries information about the velocity and gives the largest gains we observed, with lower velocity error on all eight held-out sets.

\textbf{Limitations: }Despite these improvements, the proposed approach still depends on learned priors and may degrade in scenarios that deviate significantly from the training distribution. On small training sets, such as the six flights per fold of TII-RATM, neither model generalizes across tracks, and the lower velocity error of Ex-IMO does not always translate into a lower position error. The rotor-derived torque requires rotor signals and a per-platform fit, and the public datasets were recorded indoors: in wind, airspeed and ground speed differ, which may weaken the velocity information it carries. Explicitly incorporating torque dynamics also introduces sensitivity to inaccuracies in the inertial parameters of the quadrotor, which we reduce by using only the inertia ratios. Finally, the body-frame update makes the heading appear observable through the current-estimate Jacobian, which can make the filter over-confident in yaw.

\textbf{Future work: }To further enhance generalization, future work will focus on training the model to estimate velocities by encoding drone rotational dynamics and leveraging neural system identification, on modeling the aerodynamic origin of the rotor-torque information explicitly, and on consistent (first-estimate) Jacobians for the body-frame update.

\section{CONCLUSION}

This work presented Ex-IMO, a learned inertial odometry framework for quadrotors that improves trajectory generalization by combining a body-frame representation with an extended dynamics model incorporating body-frame torque dynamics. The system integrates a Temporal Convolutional Network predicting short-horizon relative displacements with an EKF that propagates IMU measurements. Experimental results, including a factorial ablation, show that the body-frame formulation accounts for most of the improvement on previously unseen trajectories, while torque dynamics provide an additional complementary gain. An additional evaluation on four public quadrotor datasets, with trajectory shapes held out of training, confirms a lower velocity error on every held-out set and shows that the torque information is most useful when it is derived from the rotor signals. These results suggest that expressing the learning problem in physically meaningful coordinates, and then enriching it with rotational dynamics, can improve robustness and generalization in IMU-only state estimation. Future work will explore joint learning of velocity and inertial parameters to further enhance performance in highly dynamic flight regimes.

\bibliographystyle{ieeetr}
\bibliography{citation}
\end{document}